\documentclass[review]{elsarticle}
\usepackage{subfig}
\usepackage{lineno,hyperref}
\usepackage{amsmath}
\usepackage{amsfonts}
\usepackage{multirow}
\usepackage{url}
\usepackage{bm}
\modulolinenumbers[5]
\usepackage[ruled]{algorithm2e} 
\usepackage{rotating}
\newtheorem{theorem}{Theorem}
\usepackage[toc,page]{appendix}

\journal{Journal of \LaTeX\ Templates}

\begin{document}

\begin{frontmatter}

\title{Multi-level CNN for lung nodule classification with Gaussian Process assisted hyperparameter optimization}

\author[author1,author2]{Miao Zhang}

\author[author1]{Huiqi Li\corref{mycorrespondingauthor}}
\cortext[mycorrespondingauthor]{Corresponding author}
\ead{huiqili@bit.edu.com}
\author[author2,author3]{Juan Lyu}
\author[author2]{Sai Ho Ling}
\author[author2]{Steven Su}

\address[author1]{School of Information and Electronics,Beijing Institute of Technology, Beijing, 100081, China}
\address[author2]{Faculty of Engineering and Information Technology, University of Technology Sydney (UTS), 15 Broadway, Ultimo, NSW 2007, Australia}

\address[author3]{College of Information and Communication Engineering, Harbin Engineering University, Harbin, 150001, China}


\begin{abstract}
This paper investigates lung nodule classification by using deep neural networks (DNNs). DNN has shown its superiority on several medical image processing problems, like medical image segmentation, medical image synthesis, and so on, but their performance highly dependent on the appropriate hyperparameters setting. Hyperparameter optimization in DNNs is a computationally expensive problem, where evaluating a hyperparameter configuration may take several hours or even days. Bayesian optimization  has been recently introduced for the automatically searching of optimal hyperparameter configurations of DNNs. It applies probabilistic surrogate models to approximate the validation error function of hyperparameter configurations, such as Gaussian processes, and reduce the computational complexity to a large extent. However, most existing surrogate models adopt stationary covariance functions (kernels) to measure the difference between hyperparameter points based on spatial distance without considering its spatial locations. This distance-based assumption together with the condition of constant smoothness throughout the whole hyperparameter search space clearly violate the property that the points far away from optimal points usually get similarly poor performance even though each two of them have huge spatial distance between them. In this paper, a non-stationary kernel is proposed which allows the surrogate model to adapt to functions whose smoothness varies with the spatial location of inputs, and a multi-level convolutional neural network (ML-CNN) is built for lung nodule classification whose hyperparameter configuration is optimized by using the proposed non-stationary kernel based Gaussian surrogate model. Our algorithm searches the surrogate for optimal setting via hyperparameter importance based evolutionary strategy, and the experiments demonstrate our algorithm outperforms manual tuning and well-established hyperparameter optimization methods such as Random search,  Gaussian processes (GP) with stationary kernels, and recently proposed Hyperparameter Optimization via RBF and Dynamic coordinate search (HORD).

\end{abstract}


\end{frontmatter}

\section{Introduction}

Lung cancer is a notoriously aggressive cancer with sufferers having an average 5-year survival rate 18\% and a mean survival time of less than 12 months \cite{Siegel2017Colorectal}, and early diagnosis is very important to improve the survival rate. Recently, deep learning has shown its supriority in computer vision \cite{6909475,Krizhevsky2012ImageNet,Naiyan2015Transferring}, and more researchers try to diagnose lung cancers with deep neural networks to assist the early diagnosis as Computer Aided Diagnosis (CAD) systems \cite{Anthimopoulos2016Lung,Golan2016Lung,Song2017Using}. In our previous works \cite{lyvjuan}, a multi-level convolutional neural networks (ML-CNN) is proposed to handle lung nodule malignancy classification, which extracts multi-scale features through different convolutional kernel sizes. Our ML-CNN achieves state-of-art accuracies both in binary and ternary classification (which achieves 92.21\% and 84.81\%, respectively) without any preprocessing. However, the experiments also demonstrate the performance is very sensitive to hyperparameter configuration, especially the number of feature maps in every convolutional layer, where we obtain the near-optimal hyperparameter configuration through trial and error.

Automatically hyperparameter optimization is very crucial to apply deep learning algorithms in practice, and several methods including Grid search \cite{Lecun1998Efficient}, Random search \cite{Bergstra2012Random}, Tree-structured Parzen Estimator Approach (TPE)\cite{Bergstra2011Algorithms} and Bayesian optimization\cite{Snoek2012Practical}, have shown their superiority than manual search method in hyperparameters optimization of deep neural network. Hyperparameter optimization in deep neural networks is a global optimization with black-box and expensive function, where evaluating a hyperparameter choice may cost several hours or even days. It is a computational expensive problem, and a popular solution is to employ probabilistic surrogate, such as Gaussian processes (GP) and Tree-structured Parzen Estimator (TPE), to approximate the expensive error function to guide the optimization process. A stationary covariance function (kernel) is usually used in these surrogates, which depends only on the spatial distance of two hyperparameter configurations, but not on the hyperparameters themselves. Such covariance function that employs constant smoothness throughout the hyperparameter search space clearly violates the intuition that most points away from optimal point all get similarly poor performance even though each two of them have large spatial distance.

In this paper, the deep neural network for lung nodule classification is built based on multi-level convolutional neural networks, which designs three levels of CNNs with same structure but different convolutional kernel sizes to extract multi-scale features of input with variable nodule sizes and morphologies. Then the hyperparameter optimization in deep convolutional neural network is formulated as an expensive optimization problem, and a Gaussian surrogate model based on non-stationary kernel is built to approximate the error function of hyperparameter configurations, which allows the model to adapt to functions whose smoothness varies with the inputs. Our algorithm searches the surrogate via hyperparameter importance based evolutionary strategy and could find the near-optimal hyperparameter setting in limited function evaluations.

We name our algorithm as \textbf{H}yperparameter \textbf{O}ptimization with s\textbf{U}rrogate-a\textbf{S}sisted \textbf{E}volutionary \textbf{S}trategy, or HOUSES for short. We have compared our algorithm against several different well-established hyperparameter optimization algorithms, including Random search, Gaussian Process with stationary kernels, and Hyperparameter Optimization via RBF and Dynamic coordinate search (HORD) \cite{Ilija}. The main contribution of our paper is summarized as fourfold:
\begin{enumerate}[(1)]
\item A multi-level convolutional neural network is adopted for lung nodule malignancy classification, whose hyperparameter optimization is formulated as a computational expensive optimization problem. 

\item A surrogate-assisted evolutionary strategy is introduced as the framework to solve the hyperparameter optimization for ML-CNN, which utilizes a hyperparameter importance based mutation as sampling method for efficient candidate points generation. 

\item A non-stationary kernel is proposed as covariance function to define the relationship between different hyperparameter configurations, which allows the model adapt spatial dependence structure to vary with a function of location. Different with a constant smoothness throughout the whole sampling region, our non-stationary GP regression model could satisfy the assumption that the correlation function is no longer dependent on distance only, but also dependent on their relative locations to the optimal point. An input-warping method is also adopted which makes covariance functions more sensitive near the hyperparameter optimums.

\item Extensive experiments illustrate the superiority of our proposed HOSUE for hyperparaneter optimization of deep neural networks.
\end{enumerate}

We organise this paper as follows: Section II introduces the background about lung nodule classification, hyperparameter optimization in deep neural network and surrogate-assisted evolutionary algorithm. Section III describes the proposed non-stationary covariance function for hyperparameter optimization in deep neural network and the framework and details of \textbf{H}yperparameter \textbf{O}ptimization with s\textbf{U}rrogate-a\textbf{S}sisted \textbf{E}volutionary \textbf{S}trategy (HOUSES) for ML-CNN. The experimental design is described in Section IV , and we demonstrates the experimental results with discussions for state-of-the-art hyperparameter optimization approaches in Section V. We conclude and describe the future work in Section VI.

\section{Relate Works}

\subsection{Lung Nodule Classification with deep neural network}

Deep neural networks have shown their superiority to conventional algorithms in the application of computer vision, and more researchers try to employ DNNs to medical imaging diagnosis areas. Paper \cite{Sun2016Computer} presents different deep structure algorithms in lung cancer diagnosis, including stacked denoising autoencoder, deep belief network, and convolutional neural network,who obtain the binary classification accuracies 79.76\%, 81.19\% and 79.29\%, respectively. Shen et al. \cite{Shen2015Multi} proposed a Multi-scale Convolutional Neural Networks (MCNN), that utilized multi-scale nodule patches to sufficiently quantify nodule characteristics, which obtained binary classification accuracy of 86.84\%. In MCNN, three CNNs that took different nodule as inputs were assembled in parallel, and concatenated the output of each fully-connected layers as its resulting output. The experiments had shown that multi-scale inputs could help CNN learn a set of discriminative features. In 2017, they extended their research and proposed a multi-crop CNN (MC-CNN) \cite{Shen2017Multi} which automatically extracted nodule fetures by adopting a multi-crop pooling strategy, and obtained  87.14\% binary classification and 62.46\% ternary classification accuracy. In our previous works \cite{lyvjuan}, a multi-level convolutional neural networks (ML-CNN) is proposed which extracts multi-scale features through different convolutional kernel sizes. It also designs three CNNs with same structure but different convolutional kernel sizes to extract multi-scale features with variable nodule sizes and morphologies. Our ML-CNN achieves state-of-art accuracies both in binary and ternary classification (which achieves 92.21\% and 84.81\%, respectively), without any additional hand-craft preprocessing. Even though these deep learning methods were end-to-end machine learning architectures and had shown their superiority than conventional methods, the structure design and hyperparameter configuration are based on human expert’s experience through trial and error search guided by human's intuition, which is a difficult and time consuming task \cite{Negrinho2017DeepArchitect,Dong2018A}.

\subsection{Hyperparameter optimization in DNN}

Determining appropriate values of hyperparameters of DNN is a frustratingly difficult task where all feasible hyperparameter configurations form a huge space, from which we need to choose the optimal case. Setting correct hyperparameters is often critical for reaching the full potential of the deep neural network chosen or designed, otherwise it may severely hamper the performance of deep neural networks.

Hyperparameter optimization in DNN is a global optimization to find a $D$-dimensional hyperparameter setting \emph{x} that minimize the validation error \emph{f} of a DNN with learned parameters $\theta$. The optimal \emph{x} could be obtained through optimizing \emph{f} as follows:

\begin{equation} \label{[1]}
\begin{aligned}
& \min_{x\subseteq \mathbb{R}^{D}}\quad f(x,\theta;\emph{Z}_{val}) \\
& s.t.\quad \theta=\arg\min_{\theta}\ f(x,\theta;\emph{Z}_{train})
\end{aligned}
\end{equation}
where ${\emph{Z}}_{train}$ and ${\emph{Z}}_{val}$ are training and validation datasets respectively. Solving Eq.(\ref{[1]}) is very challenging for the high complexity of the function \emph{f}, and it is usually accomplished manually in the deep learning community, which largely depends on expert’s experience or intuition. It is also hard to reproduce similar results when this configuration is applied on different datasets or problems.

There are several systematic approach to tune hyperparameters in machine learning community, like Grid search, Random search, Bayesian optimization methods, and so on. Grid search is the most common strategy in hyper-parameter optimization \cite{Lecun1998Efficient}, and it is simple to implement with parallelization, which makes it reliable in low dimensional spaces (e.g., 1-$d$, 2-$d$). However, Grid search suffer from the curse of dimensionality because the search space grows exponentially with the number of hyper-parameters. Random search \cite{Bergstra2012Random} proposes to randomly sample points from the hyperparameter configuration space. Although this approach looks simple, but it could find comparable hyperparameter configuration to grid search with less computation time. Hyperparameter optimization in deep neural networks is a computational expensive problem where evaluating a hyperparameter choice may cost several hours or even days. This property also makes it unrealistic to sample many enough points to be evaluated in Grid and Random search. One popular approach is using efficient surrogates to approximate the computationally expensive fitness functions to guide the optimization process. Bayesian optimization \cite{Snoek2012Practical} built a probabilistic Gaussian model surrogate to estimate the distribution of computationally expensive validation errors. Hyperparameter configuration space is usually modeled smoothly, which means that knowing the quality of certain points might help infer the quality of their nearby points, and Bayesian optimization \cite{Bergstra2011Algorithms,Shahriari2015Taking,Bergstra2012Making} utilizes the above smoothness assumption to assist the search of hyperparameters. Gaussian Process is the most common method for modeling loss functions in Bayesian optimization for it is simple and flexible. There are several acquistion functions to determin the next promising points in Gaussian process, including Probability of Improvement (PI), Expected Improvement (EI), Upper Confidence Bound (UCB) and the Predictive Entropy Search (PES) \cite{Snoek2012Practical,Hoffman2014Predictive}.

\subsection{Surrogate-assisted evolutionary algorithm}

Surrogate-assisted evolutionary algorithm was designed to solve expensive optimization problems whose fitness function is highly computationally expensive \cite{Jin2011Surrogate,Jin2009A,Douguet2010e}. It usually utilizes  computationally efficient models, also called as surrogates, to approximate the fitness function. The surrogate model is built as:

\begin{equation} \label{[2]}
\hat{f}(x)=f^{*}(x)+\xi (x)
\end{equation}
where $f^{*}$ is the true fitness value, $\hat{f}$ is the approximated fitness value, and $\xi$ is the error function that is to minimized by the selected surrogate. Surrogate-assisted evolutionary algorithm uses one or several surrogate models $\hat{f}$  to approximate true fitness value $f^{*}$ and uses the computationally cheap surrogate to guide the search process \cite{Zhang2010Expensive}. The iteration of the surrogate-assisted evolutionary algorithm is described as: 1) Learn surrogate model $f^{*}$ based on previously truly evaluated points $(x, f(x))$; 2) Utilize $f^{*}$ to evaluate new mutation-generated points and find the most promising individual $x^{*}$. 3) evaluate the true fitness value of additional points$(x^{*}, f(x^{*}))$ . 4) Update training set.

Gaussian process, polynomials, Radial Basis Functions (RBFs),  neural networks, and Support Vector Machines are major techniques to approximate true objective function for surrogate model learning.  A non-stationary covariance function based Gaussian process is adopted as the surrogate model in this paper, which allows the model adapt spatial dependence structure to vary with locations and satisfies our assumption that the hyperparameter configuratio performs well near the optimal points while poorly away from the optimal point. Then the evolutionary strategy is used to search the near-optimal hyperparameter configuration. The next section will present the details of our Hyperparameter Optimization with sUrrogate-aSsisted Evolutionary Strategy (HOUSES) for ML-CNN.

\section{\textbf{H}yperparameter \textbf{O}ptimization with s\textbf{U}rrogate-a\textbf{S}sisted \textbf{E}volutionary \textbf{S}trategy}
\begin{figure*}[tb]
\includegraphics[width=12cm,height=5cm]{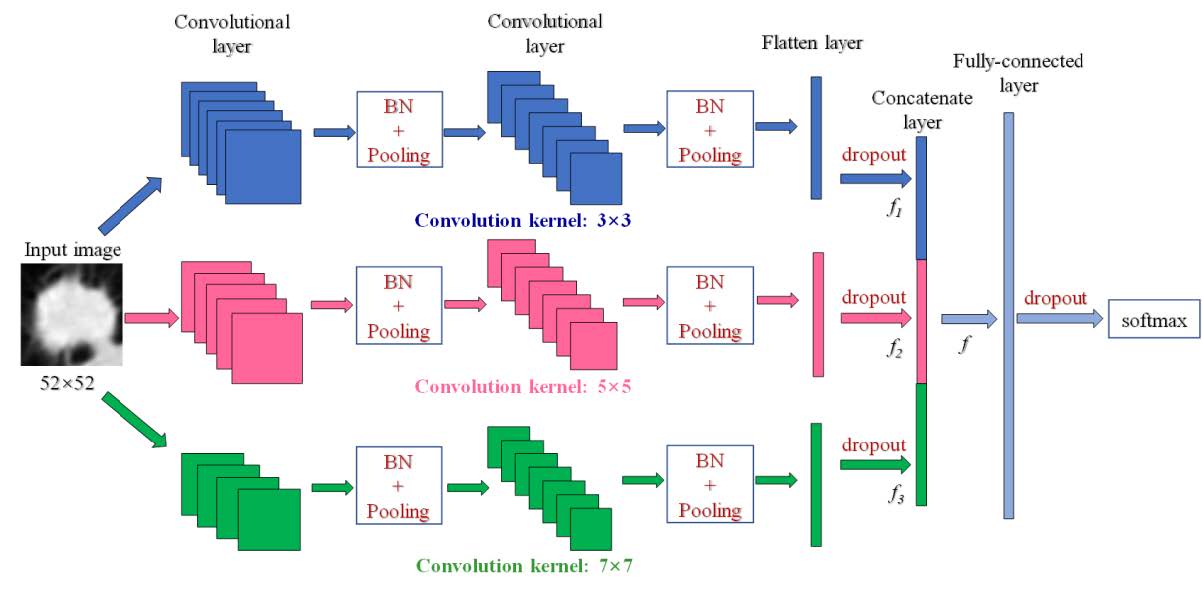}
\caption{The structure of proposed ML-CNN for lung nodule malignancy classification \cite{lyvjuan}.}
\label{figure1}
\end{figure*}

In our previous work \cite{lyvjuan}, a multi-level convolutional neural network is proposed for lung nodules classification, which applies different kernel sizes in three parallel levels of CNNs to effectively extract different features of each lung nodule with different sizes and various morphologies. Fig. \ref{figure1} presents the structure of ML-CNN, which contains three level of CNNs and each of them has same structure and different kernel size. As suggested in our previous work, feature maps number in each convolutional layer has significant impact on the performance of ML-CNN, so as the dropout rates. The hyperparameter configuration of ML-CNN in  \cite{lyvjuan} is based on trial and error manual search approach, which is a time-consuming work for researcher and has no guarantee to get an optimal configuration. In this section, we introduce \textbf{H}yperparameter \textbf{O}ptimization with s\textbf{U}rrogate-a\textbf{S}sisted \textbf{E}volutionary \textbf{S}trategy(HOUSES) to our ML-CNN for lung nodule classification, which could automatically find a competitive or even better hyperparameter configuration than manual search method without too much computational cost. The framework of the proposed HOUSES for ML-CNN is presented in \textbf{Algorithm 1}. In our hyperparameter optimization method, a non-stationary kernel is proposed as covariance function to define the relationship between different hyperparameter configurations, which allows the model adapt spatial dependence structure to vary with a function of location, and the algorithm searches for the most promising hyperparameter values based on surrogate model through evolutionary strategy. In our HOUSES, several initial hyperparameter configuration points are randomly generated through Latin Hypercube Sampling (LHS) \cite{Iman2008Latin} methods to keep diversity of the initial population. These initial points are truly evaluated and used as the training set $Tr_0\{(\textbf{x}_i,f_i)\}_{i=1}^{n_0}$ to build the initial surrogate model. Then the evolutionary strategy generates a group new points which are evaluated according to the acquisition function of the surrogate model. Several most promising individuals $x^{*}$ are found from those new generated points based on acquisition function and then truly evaluated. The most promising points with true fitness value$(x^{*}, f(x^{*}))$ are added to training set to update surrogate model. We describe our HOUSES in the following paragraphs.

\begin{algorithm}
\label{algorithm1}
\caption{General Framework of HOUSES}
\KwIn{Initial population size $\bm{n_0}$, Maximum generation $\bm{g_{max}}$, Mutation rate $p_m$, number of new generated points every generation $\bm{m}$, $Dataset$, $DNN\ model$}
\KwOut{best hyperparameter configuration $c_{best}$ for DNN model}
\textbf{Divide} dataset into Training, Validation and Testing sets

\textbf{Initialization} A hyperparameter configuration population $pop_{0}$ is randomly generated through Latin Hypercube Sampling. These hyperparameter points are used to train DNN model in Training set, and truly evaluated in the Validation set to get true fitness values $Tr_0=\{(\textbf{x}_i,f_i)\}_{i=1}^{n_0}$.

\While{Maximum generation $g_{max}$ is not reached}{
\textbf{1.} Use $Tr$ to fit or update the Gaussian surrogate model $\hat{f}$ according Eq.\eqref{[3]};

\textbf{2.} $\boldsymbol{pop_{selected}}$= \textbf{select} ($\boldsymbol{pop_{g}}$)// \emph{select individuals with good performance and diversity for mutation};

\textbf{3.} $\boldsymbol{pop_m}$= \textbf{mutation} ($\boldsymbol{pop_{selected}}$)// \emph{apply mutation operation to selected points to generate $\bm{m}$ new points};

\textbf{4.} Calculate $\{(\textbf{x}_i,\hat{f}_i)\}_{i=1}^{\bm{m}}$ for $\bm{m}$ new generated points based on Gaussian surrogate model and acquisition functions Eq.\eqref{[11]}\eqref{[12]}\eqref{[13]};

\textbf{5.} Set $\mathbf{x^*}=\textbf{argmin}{\{\hat{f}_i\}_{i=1}^{\bm{m}}}$;

\textbf{6.} Truly evaluate $f(\mathbf{x^*})$ in Training set and Validation set to get true fitness values;

\textbf{7.} Update $Tr_{g+1}=\{Tr_{g+1}\cup (\mathbf{x^*},f(\mathbf{x^*}))\}$;
}

\textbf{Return} the hyperparameter configuration $c_{best}$.

\end{algorithm}

\subsection{surrogate model building}

Gaussian process (also known as Kriging) is choosed as the surrogate model in HOUSES searching for the most promising hyperparameters, which uses a generalization of the Gaussian disribution to describe a function, defined by a mean $\mu$, and covariance function $\sigma$:

\begin{equation} \label{[3]}
\hat{f}(x)\sim \boldsymbol{N}(\bm{\mu}(x),\sigma(x))
\end{equation}

Given training data that consists $n$ $D$-dimensional  inputs and outputs, $\{x_{1:n},f_{1:n}\}$, where $x_i\subseteq \mathbb{R}^{D}$ and  $f_{i}=f(x_i)$. The predictive distribution based on Gaussian process at an unknown input, $x^{*}$, is calculated by the following:

\begin{equation} \label{[4]}
\mu(x^{*})=K_{*}(K+\theta_{c}^{2}I)^{-1}f_{i:n}
\end{equation}
\begin{equation} \label{[5]}
\sigma(x^{*})=K_{**}-K_{*}((K+\theta_{c}^{2}I)^{-1})K_{*}^{T}
\end{equation}
where $K_{*}=[k(x^{*},x_{1}),...,k(x^{*},x_{n})]$ and $K_{**}=k(x^{*},x^{*})$, $\theta_{c}$ is a noise parameter, $K$ is the associated covariance matrix which is built as:
\begin{equation} \label{[6]}
K=\begin{bmatrix}
k(x_{1},x_{1}) &\hdots  &k(x_{1},x_{n})\\ 
 \vdots& \ddots & \vdots\\ 
 k(x_{n},x_{1})& \hdots & k(x_{n},x_{n})
\end{bmatrix}
\end{equation}
$k$ is a covariance function that defines the relationship between points in the forms of a kernel. A used kernel is automatic relevance determination (ARD) squared exponential covariance function:
\begin{equation} \label{[7]}
k(x_{i},x_{j})=\theta_{f}\ \textbf{exp}\sum_{d=1}^{D}\frac{-(x_{i}^{d}-x_{j}^{d})^{2}}{2\theta_{d}^{2}} 
\end{equation}

\subsection{Non-stationary covariance function for hyperparameter optimization in DNNs}

\subsubsection{Spatial location transformation}

In the hyperparameter optimization of DNNs, two far away hyperparameter points usually perform both poorly when they are away  optimal point. This property means that the correlation of two hyperparameter configuration depends not only on the distance between them, but also the points' spatial locations. Those stationary kernel, such as Gaussian kernel, clearly could not satisfy this property of hyperparameter optimization in DNNs. To account for this non-stationarity, we proposed a non-stationary covariance function, where we use the relative distance to optimal point to measure the spatial location difference of two hyperparameter points. The relative distance based kernel is defined as:
\begin{equation} \label{[8]}
k({x_{i},x_{j}})=\theta_{f} \textbf{exp}\sum_{d=1}^{D}\frac{-(\left |x_{i}^{d}-\boldsymbol{s}^{d}  \right |-\left |x_{j}^{d}-\boldsymbol{s}^{d}  \right |)^{2}}{2\theta_{d}^{2}} 
\end{equation}
where $\boldsymbol{s}$ is the assumed optimal point. It is also easy to prove this relative distance based covariance function $k({x_{i},x_{j}})$ is a kernel based on \textbf{Theorem 1} \ref{the1}. Eq.\eqref{[8]} could be obtained by set $\psi(\textbf{x})=\left | \textbf{x}-\textbf{s} \right |$ and $k'$ as Gaussian kernel. This relative distance based kernel is no longer a function of distance between two points, but depends on their own spatial locations to the optimal point. 

\begin{theorem}\label{the1}
 if $\psi$ is an $\mathbb{R}^{D}$-valued function on $\boldsymbol{X}$ and $k'$ is a kernel on  $\mathbb{R}^{D}\times \mathbb{R}^{D}$, then
\begin{equation} \label{[9]}
k(\textbf{x},\textbf{z})=k'(\psi(\textbf{x}),\psi(\textbf{z}))
\end{equation}
is also a kernel.

\textbf{Proof}: $k': \mathbb{R}^{D}\times \mathbb{R}^{D}\rightarrow \mathbb{R}$, $\psi: \mathbb{R}^{D}\rightarrow \mathbb{R}^{D}$, $k'$ is a valid kernel, then we have
\begin{equation} \nonumber
k'(\textbf{x},\textbf{z})=\varphi (\textbf{x})^{\textbf{T}}\varphi (\textbf{z})
\end{equation}
so that 
\begin{equation}\nonumber
k(\textbf{x},\textbf{z})=\varphi(\psi(\textbf{x}))^{\textbf{T}}\varphi (\psi(\textbf{z}))
\end{equation}
is a kernel.

\end{theorem}

\subsubsection{Input Warping}

In the hyperparameter optimization of machine learning models, objective functions are usually more sensitive near the optimal hypeparameter setting while much less sensitive far away from the optimum. For example, if the optimal learning rate is 0.05, it is supposed to obtain 50\% performance increase when the learning rate changing from 0.04 to 0.05, while may just 5\% increase from 0.25 to 0.24. Traditionally, most researchers often use logarithm function to transform the input space and then search in the transformed space, which is effective only when the non-stationary property of the input space is known in advance. Recently, a beta cumulative distribution function is proposed as the input warping transformation function  \cite{snoek2014input,swersky2017improving},
\begin{equation} \label{[10]}
w_d(\textbf{x}_d)=\int_{0}^{\textbf{x}_d}\frac{u^{\alpha _d-1}(1-u)^{\beta _d-1}}{B(\alpha _d,\beta _d)}\textbf{d}u
\end{equation}
where $B(\alpha _d,\beta _d)$ is the beta function, which is to adjust the shape of input warping function to the original data based on parameters $\alpha _d$, and $\beta _d$.

\begin{figure*}[tb]
\centering
\includegraphics[width=13cm,height=4cm]{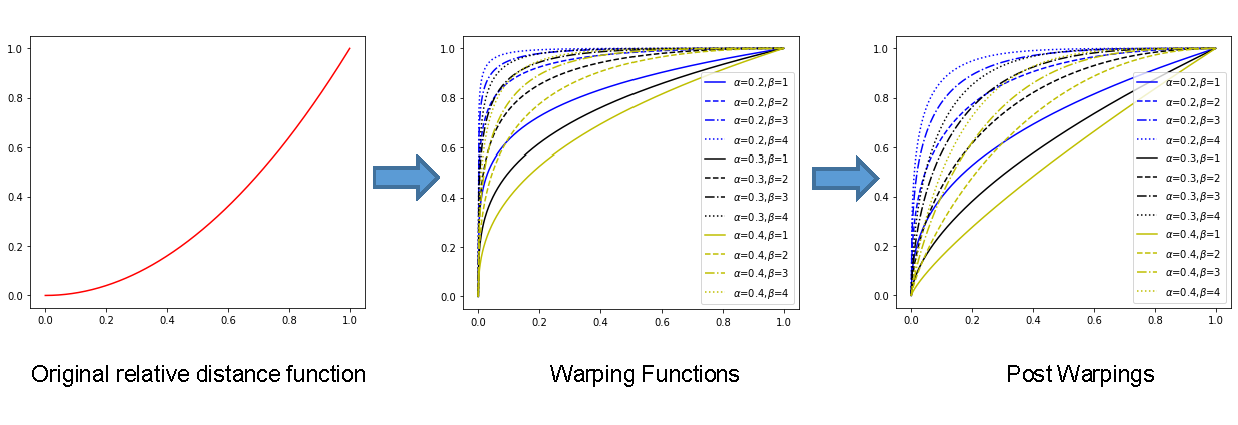}
\caption{Example of how Kumaraswamy cumulative distribution function transforming a concave function into a convex function, which makes the kernel function is much more sensitive to small inputs.}
\label{figure2}
\end{figure*}

Different from \cite{snoek2014input,swersky2017improving}, we just take the relative distance to local optimum as inputs to be warped that make kernel function is more sensitive to small inputs and less sensitive for large ones. We take the Kumaraswamy cumulative distribution function as the substitute, which is not only because of computational reasons, but also it is easier to fulfill the non-stationary property of our kernel function after spatial location transformation,

\begin{equation} \label{[11]}
w_d(\textbf{x}_d)=1-(1-\textbf{x}_d^{\alpha _d})^{\beta _d}
\end{equation}

Similar to Eq.\eqref{[9]}, it is easy to be proven  that $k(\textbf{x},\textbf{x}')=k'(w(\psi(\textbf{x})),w(\psi(\textbf{x}')))$ is a kernel. Fig.\ref{figure2} illustrates input warping example with different shape parameters $\alpha _d$, and $\beta _d$ input warping functions. And the final kernel for our HOUSE is defined as: 

\begin{equation} \label{[12]}
k({x_{i},x_{j}})=\theta_{f}\ \textbf{exp}\sum_{d=1}^{D}\frac{-(w_d\left |x_{i}^{d}-\boldsymbol{s}^{d}  \right |-w_d\left |x_{j}^{d}-\boldsymbol{s}^{d}  \right |)^{2}}{2\theta_{d}^{2}} +\theta_{k}\ \textbf{exp}\sum_{d=1}^{D}\frac{-(w_d\left |x_{i}^{d}-x_{j}^{d} \right |)^{2}}{2\gamma_{d}^{2}}
\end{equation}
Eq.\eqref{[12]} is also a kernel proved based on \textbf{Theorem 2} \ref{the2}. This non-stationary kernel Eq.\eqref{[12]} satisfies the assumption that the correlation function of two hyperparameter configuration is not only dependent on their distances, but their relative locations to optimal point. However it is impossible to get the optimal point in advance, and we instead use the hyperparameter configuration with best performance in the train set, and updates it in every iteration in our proposed HOUSES.

\begin{theorem}\label{the2}
If $k_1$ is a kernel on  $\mathbb{R}^{D}\times \mathbb{R}^{D}$ and $k_2$ is also a kernel on  $\mathbb{R}^{D}\times \mathbb{R}^{D}$, then
\begin{equation} \label{[13]}
k(\textbf{x},\textbf{z})=k_1(\textbf{x},\textbf{z})+k_2(\textbf{x},\textbf{z})
\end{equation}
is also a kernel.

\textbf{Proof}: This is because if $k_1(\textbf{x},\textbf{z})$ and $k_2(\textbf{x},\textbf{z})$ are valid kernels on  $ \mathbb{R}^{D}\times \mathbb{R}^{D}\rightarrow \mathbb{R}$, then we have $k_1(\textbf{x},\textbf{z})=\varphi^{\textbf{T}}(\textbf{x})\varphi(\textbf{z})$ and $k_2(\textbf{x},\textbf{z})=\psi^{\textbf{T}}(\textbf{x})\psi(\textbf{z})$, we may define 
\begin{equation} \nonumber
\theta (\textbf{x})=\varphi(\textbf{x})\oplus \psi(\textbf{x})=[\varphi(\textbf{x}), \psi(\textbf{x})]^\textbf{T}
\end{equation}
so that
\begin{equation} \nonumber
k(\textbf{x},\textbf{z})=\theta(\textbf{x})^{\textbf{T}}\theta(\textbf{z})
\end{equation}
is a kernel.

\end{theorem}

\subsection{Acquisition function}

After building a surrogate model, an acquisition function is required to choose the most promising point for truly evaluation. Different with surrogate model that approximates the optimizing problem, the acquisition function is to be utilized to find the most possible optimal solution. 

 We applied three different acquisition functions for Gaussian process (GP) based hyperparemeter optimization:
\begin{itemize}
\item Probability of Improvement
\begin{equation} \label{[14]}
\alpha _{\mathbf{PI}}(\mathbf{x} )=\Phi (\gamma (\mathbf{x})), \qquad\qquad  \gamma (\mathbf{x})=\frac{f(\mathbf{x_{\mathbf{best}}})-\mu(\mathbf{x})}{\sigma (\mathbf{x})}
\end{equation}
where $\Phi (z)=(2\pi)^{-\frac{1}{2}}\int_{z}^{-\infty}\textbf{exp}(\frac{-t^2}{2})dt$.

\item Expected Improvement
\begin{equation} \label{[15]}
\alpha _{\mathbf{EI}}(\mathbf{x} )=\sigma (\mathbf{x})(\gamma (\mathbf{x})\Phi (\gamma (\mathbf{x}))+\boldsymbol{N}(\mu(x)))
\end{equation}
where $\boldsymbol{N}(z)$  is  the variable $z$ has a Gaussian  distribution with $z\sim \boldsymbol{N}(0,1)$.

\item and Upper Confidence Bound
\begin{equation} \label{[16]}
\alpha _{\mathbf{UCB}}(\mathbf{x} )=\mu(\mathbf{x})+\emph{w}\sigma (\mathbf{x})
\end{equation}
with a tunable $\emph{w}$ to balance exploitation against exploration \cite{Ilija}.
\end{itemize}

\subsection{Hyperparameter Importance based Mutation for candidate hyperparameter points generation}

The mutation aims to generate better individuals through mutating selected excellent individuals, which is a key step of optimization in evolutionary strategy. To maintain the diversity of the population, a uniformed selection strategy is adopted in mutation. It first divides every dimension into $M$ uniformed grids \cite{Zhang2018A}, and the point with the highest fitness in every dimensional grid is selected for mutation. In this way, $D*M$ individules are selected and the polynomial mutation is applied to every selected individual to generate $n_d$ candidate hyperparameter points, respectively. These $D*M*n_d$ points are evaluated based on acquisition fuction, and the most promising point is selected for truly evaluation and added into the training set to update surrogate model.

\begin{figure*}
  \centering
  \begin{minipage}{6cm}
      \includegraphics[width=6cm]{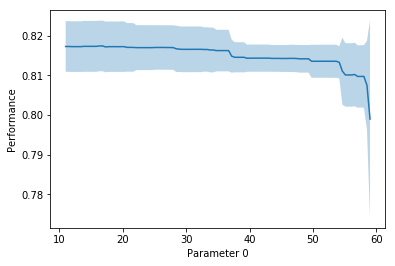}
  \end{minipage}
    \begin{minipage}{6cm}
      \includegraphics[width=6cm]{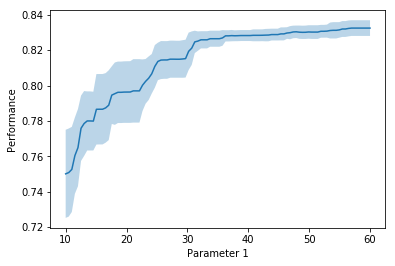}
  \end{minipage}
  
  \begin{minipage}{6cm}
      \includegraphics[width=6cm]{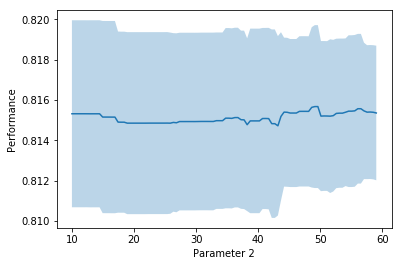}
  \end{minipage} 
  \centering
  \begin{minipage}{6cm}
      \includegraphics[width=6cm]{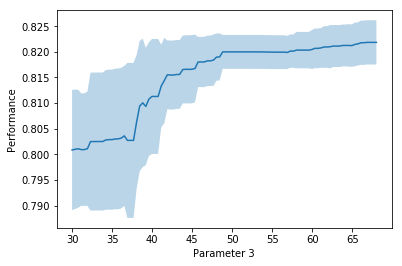}
  \end{minipage}
  \begin{minipage}{6cm}
      \includegraphics[width=6cm]{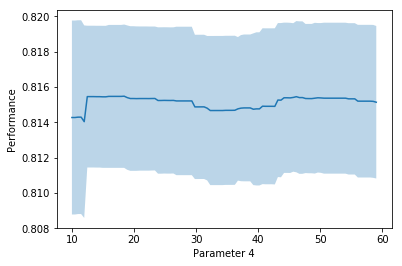}
  \end{minipage}  
   \begin{minipage}{6cm}
      \includegraphics[width=6cm]{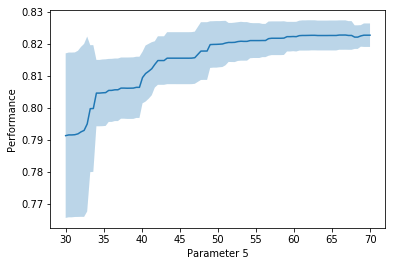}
  \end{minipage}

  \caption{Functional ANOVA based marginal response performance of the number of feature maps of all convolutional layers in three different levels of  ML-CNN. The first two parameters are for the two convolutional layers in the first level, the middle two are for the second level, and the last two are for the third level. Results show that the latter ones in the three level brings more effects to the performance, while there is no significant difference among all possible configuration for the previous feature maps number in each level of ML-CNN.}
 \label{figure3}
\end{figure*}

However, as suggested by several recent works on Bayesian based hyperparameter optimization \cite{hoos2014efficient,Bergstra2012Random}, most hyperparameters are truly unimportant while some hyperparameters are much more important than others. Fig. \ref{figure3} demonstrates ML-CNN marginal performance variation with the number of feature maps, which clearly shows that the number of feature maps in the last convolutioanry layer in every layer is much more crucial to ML-CNN than previous ones. Paper \cite{hoos2014efficient} proposed to use Functional analysis of variance (functional ANOVA) to measure the importance of hyperparameters in machine learning problems. Functional ANOVA is a statistical method for prominent data analysis, which partitions the observed variation of a response value (CNN performance) into components due to each of its inputs (hyperparameter setting). Function ANOVA is able to illustrate how the response performance changes with input hyperparameters. It first accumulates the response function values of all subsets of it inputs $N$:

\begin{equation} \label{[17]}
\hat{y}(\theta )=\sum_{U\subseteq N}\hat{f}_U(\theta_U)
\end{equation} 
where the component $\hat{f}_U(\theta_U)$ is defined as:

\begin{equation} \label{[18]}
\hat{f}_U(\theta_U)=\left\{\begin{matrix}
\hat{f}_{\emptyset} &\mathbf{if}\  U=\emptyset   \\ 
\hat{a}_U(\theta_U)-\sum \hat{f}_W(\theta_W)& \mathbf{otherwise}
\end{matrix}\right.
\end{equation} 
where the constant $\hat{f}_{\emptyset}$ is the mean value of the function over its domain, $\hat{a}_U(\theta_U)$ is the marginal predicted performance defined as $\hat{a}_U(\theta_U)=\frac{1}{\left \| \Theta_T \right \| }\int \hat{y}(\theta_{N\mid U})d\theta_T$. The subset $\left |U  \right |>1$ captures the interaction between all hyperparameters in subset $U$, while we only consider the separate hyperparameter importance in this paper and set $\left |U  \right |=1$. The component function $\hat{f}_U(\theta_U)$ is then calculated as: 
\begin{equation} \label{[19]}
\hat{f}(\theta_d)=\hat{a}(\theta_d)=\frac{1}{\left \| \Theta_T \right \|}\int \hat{y}(\theta_{N\mid d})\textbf{d}\theta_T
\end{equation} 
where $\theta_d$ is the single hyperparameter, $T=N\setminus d$, $\Theta_T=\Theta\setminus \theta_i$, $\Theta=\theta_1\times \cdots \times \theta_D$. The variance of response performance of $\hat{y}$ across its domain $\Theta$ is 
\begin{equation} \label{[20]}
\mathbb{V}=\sum _{i=1}^{n}\mathbb{V}_d,\    \mathbb{V}_d=\frac{1}{\left \| \theta_d \right \|}\int\hat{f}(\theta_i)^2\textbf{d}\theta_d
\end{equation} 
The importance of each hyperparameter could thus be quantified as:
\begin{equation} \label{[21]}
\mathbb{I}_d=\mathbb{V}_d/\mathbb{V}
\end{equation} 

When the polynomial mutation operator is applied to individuals, genes corresponding to different hypeparameters have different mutation probabilities in terms of hyperparameter importances, where genes with lager importances are supposed to have higher mutation probabilities to generate more offsprings. In this way, our evolutionary strategy is supposed to put more emphases in those subspaces of important hyperparameters and find better hyperparameter settings.

\section{Experimental Design}

To examine the optimization performance of our proposed HOUSES for hyperparameter optimization, two sets of experiments have been conducted. We test HOUSES on a Multi-layer Perceptron (MLP) network and LeNet applied to the popular MNIST dataset, and AlexNet applied to CIFAR-10 dataset in the first one. For second set, there is only one experiment whose target is to find an optimal hyperparameter configuration of ML-CNN applied to lung nodule classificayion. All the experiments were performed with Nvidia Quadro P5000 GPU (16.0 GB Memory, 8873 GFLOPS). Our experiments are implemented in Python 3.6 environment, and Tensorflow \footnote{\url{https://github.com/tensorflow/tensorflow}} and Tensorlayer \footnote{\url{https://github.com/tensorlayer/tensorlayer}} are used for building deep neural networks.

The following subsections present a brief introduction of experimental problems, peer algorithms, and evaluation budget and experimental setting.

\subsection{DNN problems}

The first DNN problem in the first experimental set is MLP network applied to MNIST, which consists of three dense layers with ReLU activation and dropout layer between them and SoftMax at the end. The hyperparameters we optimize with HOUSES and other peer algorithms include dropout rate in each dropout layer and number of units in dense layers. This problem has 5 parameters to be optimized, which is described as 5-MLP in this paper. The second DNN problem is LeNet5 applied to MNIST that has 7 hyperparameters to be optimized, which is described as 7-CNN. The 7-CNN contains two convolutional blocks, each containing one convolutional layer with batch normalization, followed by $ReLU$ activation and $2\times 2$ max-pooling, and three fully-connected layers with two dropout layers among them are followed at the end. The optimizing parameters in 7-CNN contain the feature maps number in every convolutional layer, the units number in the first two fully-connected layers and also thedropout rates in all dropout layers. The third DNN problem in the first set is to optimize the hyperparameters of AlexNet applied to CIFAR-10 dataset. There are 9 parameters: feature numbers in 5 converlutional layers, numbers of units in two fully-connected layers and the dropout rate of the dropout layer after them. This is described as 9-CNN problem in this paper.

In the second experimental set, we evaluate HOUSES on ML-CNN applied to lung nodule classificayion. There are also 9 hyperparameters to be optimized, which consists of number of feature maps in every convolutioanl layer, number of unites in full-connected layer, and dropout rate of every dropout layer. This hyperparameter optimization problem is denoted as 9-ML-CNN in this paper. The lung nodule images in this experiment are from the Lung Image Database Consortium (LIDC) and Image Database Resource Initiative (IDRI) database \cite{3Rd2012The,Reeves2007The}, containing 1,018 cases from 1,010 patients and are annotated by 4 radiologists. The malignancy suspiciousness of each nodule in the database is rated from 1 to 5 annotated by four radiologists, where level 1 and 2 are benign nodules, level 3 is indeterminate nodule and level 4 and 5 are malignant nodule. The diagnosis of nodule is labeled to the class with the highest frequency, or indeterminate when more than one class have the highest frequency. The nodules in the images are cropped according to the contour annotations of 4 radiologists and resized by $52\times 52$ as the input of our multi-level convolutional neural networks.

\subsection{Peer algorithm}
We compare HOUSES against Random search, Gaussian processes (GP) with Gaussian kernel, and Hyperparameter Optimization via RBF and Dynamic coordinate search (HORD). We also compared three different acquisition functions for Gaussian processes (GP) based hyperparemeter optimization: Gaussian processes with Expected Improvement (GP-EI), Gaussian processes with Probability of Improvement (GP-PI), and Gaussian processes with Upper Confidence Bound (GP-UCB).

\subsection{Evaluation budget and experimental setting}
Hyperparameter configuration evaluation is typically computationally expensive which consists of the most computation cost in DNN hyperparameter optimization problem. For fair comparison, we set the number of function evaluations as 200 for all comparing algorithms. The number of training iterations for MNIST dataset is set as 100, and CIFAR-10 and LIDC-IDRI are set as 200 and 500, respectively.

We implement the Random search with the open-source HyperOpt library \footnote{\url{http://hyperopt.github.io/hyperopt/}}. We use the public sklearn library \footnote{\url{https://scikit-learn.org/stable/modules/gaussian_process.html}} to build Gaussian Processes based surrogate model. The implementation for HORD is at \footnote{\url{bit.ly/hord-aaai}}. The code for hyperparameter importance assessing based on functional ANOVA is available at \footnote{\url{https://github.com/automl/fanova}}.

\section{Experimental results and discussion}

\begin{table*}
\centering
\caption{Experimental mean accuracy of comparing algorithms on 4 DNN problems}
\label{table3}
\setlength{\tabcolsep}{6pt}
\begin{tabular}
{p{80pt}|p{40pt}p{40pt}p{40pt}p{55pt}}
\hline
DNN Problems&5-MLP&7-CNN&9-CNN&9-ML-CNN\\
\hline
\hline
Random Search&0.9731&0.9947&0.7429&0.8401\\
\hline
HORD&0.9684&0.9929&0.7471&0.8421\\
\hline
GP-EI&0.9647&0.9934&0.7546&0.8517\\
\hline
GP-PI&0.9645&0.9937&0.7650&0.8473\\
\hline
GP-UCB&0.9637&0.9942&0.7318&0.8457\\
\hline
HOUSES-EI&0.9698&0.9931&0.7642&0.8511\\
\hline
HOUSES-PI&0.9690&0.9949&\textbf{0.7683}&0.8541\\
\hline
HOUSES-UCB&\textbf{0.9738}&0.9937&0.7493&\textbf{0.8576}\\
\hline
Manual Tuning&-&-&-&0.8481\\
\hline
\end{tabular}
\end{table*}

\begin{sidewaystable}[!htbp]
    \caption{Comparison results in three DNN problems}
    \label{t2}
    \begin{center}
        \begin{tabular}{c|ccc|ccc|ccc}
            \cline{1-10}
            \multirow{2}{*}{Algorithm} &\multicolumn{3}{c}{5-MLP}&\multicolumn{3}{c|}{7-CNN}&\multicolumn{3}{c}{9-CNN}\\
            \cline{2-10}            
            &Sensitivity&Specificity&AUC&Sensitivity&Specificity&AUC&Sensitivity&Specificity&AUC\\
            \hline
            \hline
            Random Search&0.95054&0.99521&0.97588&0.98809&0.99569&0.99339&0.7590&0.97130&0.8617\\
            \hline
            HORD&0.95085&0.99458&0.97272&0.98398&0.99832&0.99111&0.76690&0.97410&0.87050\\
            \hline
            GP-EI&0.95474&0.99502&0.97488&0.95576&0.99840&0.99182&0.76800&0.97420&0.87110\\
            \hline
            GP-PI&0.93838&0.99324&0.96581&0.98706&0.99857&0.99281&0.7571&0.9730&0.86505\\
            \hline
            GP-UCB&0.93604&0.99298&0.96451&0.98511&0.99842&0.99207&0.7609&0.97343&0.86717\\
            \hline
            HOUSES-EI&0.93642&0.99300&0.96472&0.98414&0.99855&0.99519&0.76940&0.97438&0.87200\\
            \hline
            HOUSES-PI&0.94486&0.99395&0.96940&0.98517&0.99857&0.99377&\textbf{0.7798}&\textbf{0.97553}&\textbf{0.87767}\\
            \hline
            HOUSES-UCB&\textbf{0.96161}&\textbf{0.99578}&\textbf{0.97870}&0.98578&0.99852&0.99355&0.7609&0.9343&0.86717\\
            \hline                      
        \end{tabular}
    \end{center}
\end{sidewaystable}

\subsection{Experiments on MNIST and CIFAR-10}
In this section, we evaluate these peer hyperparameter optimization algorithms on 3 DNN problems, including MLP applied to MNIST (5-MLP),and LeNet network to MNIST (7-CNN), and AlexNet applied to CIFAR10 (9-CNN).

For 5-MLP problem, Table 1 (Column 2) shows the obtained test results of different comparing methods, and Figure 2 (a) also plots the  average accuracy over epochs of the obtained hyperparameter configurations from different hyperparameter optimization methods. One surprised observation from the above table and figure is that the simplest Random Search method could get satisfied results, which sometimes even outperforms some Bayesian optimization based methods (GPs and HORD). This phenomenon suggests that, for low-dimensional hyperparameter optimization, the simple Random Search could perform very well, which is also in line with \cite{Bergstra2012Random}. Furthermore, we can also find from Table 1 (Column 2) and Figure 2 (a) that, with the same experimental settings, our proposed non-stationary kernel clearly perform better comparing with stationary Gaussian kernel with all three acquisition functions in 5-MLP problem. It is also demonstrates that incorporating priors based on expert intuition into Bayesian optimization and designing a non-stationary kernel is necessary for Gaussian processes based hyperparameter optimization.

In the 7-CNN problem, we found that most hyperparameter optimization algorithms could obtain satisfied result, and whose test errors are less than the best result in 5-MLP problem (see Column 3 of Table 1). These results demonstrate that a better neural network structure could significantly improve the performance, and is more robust to hyperparameter configuration, where there is not much significant difference for those hyperparameter optimization methods in 7-CNN problem. So designing an appropriate neural network structure is the first importance, and this is also the reason why we design a Multi-Level Convolutionary Neural Network for lung nodule classification.

As to the more complicated and harder DNN problem 9-CNN, GPs could found significantly better hyperparameters than Random Search algorithm, except GP-UCB, which may due to the improper weighting setting in UCB acquisition function (see Column 4 of Table 1, Figure 2 (c)).  These results also shows that Random Search algorithm performs extremely poorer than other hyperparameter optimization algorithms in 9-CNN problem, and suggests that a hyperparameter optimization is required in complicated DNN problems which helps the deep neural network to reach the full potential. Similar to those results in 5-MLP problem, the results in 9-CNN again show the superiority  of our proposed non-stationary kernel for Hyperparameter optimization in CNN, where non-stationary kernel always outperforms than standard Gaussian kernel. Figure 2 (c) shows that the performances of HOUSES and GP with different acquisition functions are distinguishable, where HOUSE-PI and GP-PI clearly get better results than other two acquisition functions. In addition, the results also show the importance of a suitable acquisition function, where GPs with UCB acquisition function even get worse results than Random Search in 9-CNN while get the best results in 5-MLP problem among all methods. 

Table 2 summarizes the sensitivity and specificity (accuracy has been present in Table 1) of hyperparameter configuration obtained by all comparing algorithm for 5-MLP, 7-CNN, and 9-CNN, and the three indicators are defined as:

\begin{equation} \label{[22]}
\begin{aligned}
& Accuracy=(TP+TN)/(TP+FP+FN+TN) \\
& Sensitive=TP/(TP+FN)\\
&Specificity=TN/(TN+FP)
\end{aligned}
\end{equation}
We also calculated Area Under Curve (AUC) \cite{Chien2003Pattern} as the assessment criteria for Receiver Operating Characteristic (ROC) curve in Table 2. As demonstrated in Table 2, our HOUSES appraoch outperforms Random Search, HORD and normal kernel based Gaussian processes in accuracy, sensitivity and specificity in 5-CNN and 9-CNN problems. In 5-MLP problem, Random Search also gets incredible results, which also suggests that the simple Random Search could perform very well in low-dimensional hyperparameter optimization. Although there are just 1-2 percentage increase on the classification rate, it is a significant improvement for hyperparameter optimization. There is no much statistic differences between these comparing algorithms in the results of 7-CNN, which again demonstrates that a better neural network structure could significant improve the performance and relieve the work of hyperparameter optimization works.

\begin{figure*}
  \centering
  \begin{minipage}{6cm}
      \includegraphics[width=6cm]{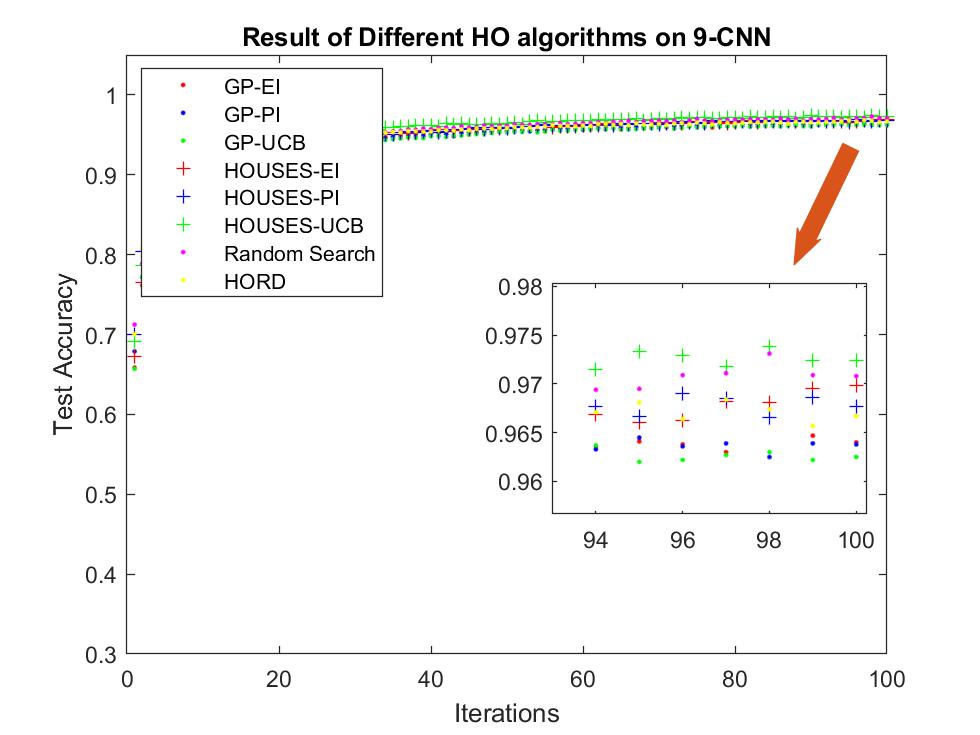}
      \caption*{(a)Testing accuracy of all hyperparameter optimization algorithms on 5-MLP problem.}
  \end{minipage}
  \begin{minipage}{6cm}
      \includegraphics[width=6cm]{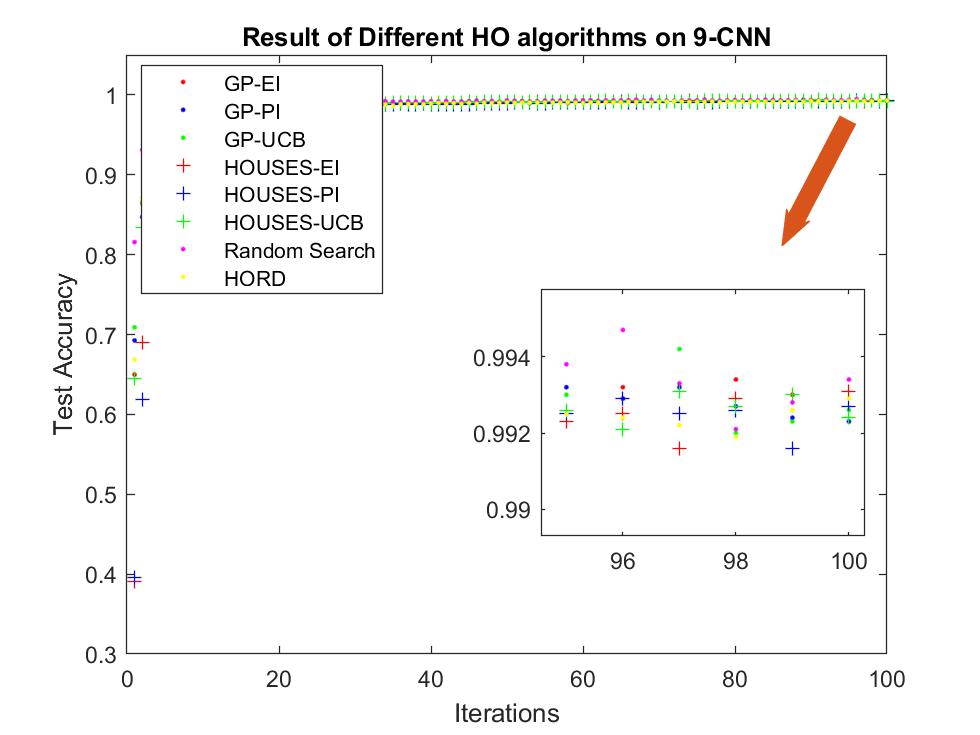}
      \caption*{\quad (b) Testing accuracy of all hyperparameter optimization algorithms on 7-CNN problem.}
  \end{minipage} 
  \centering
  \begin{minipage}{6cm}
      \includegraphics[width=6cm]{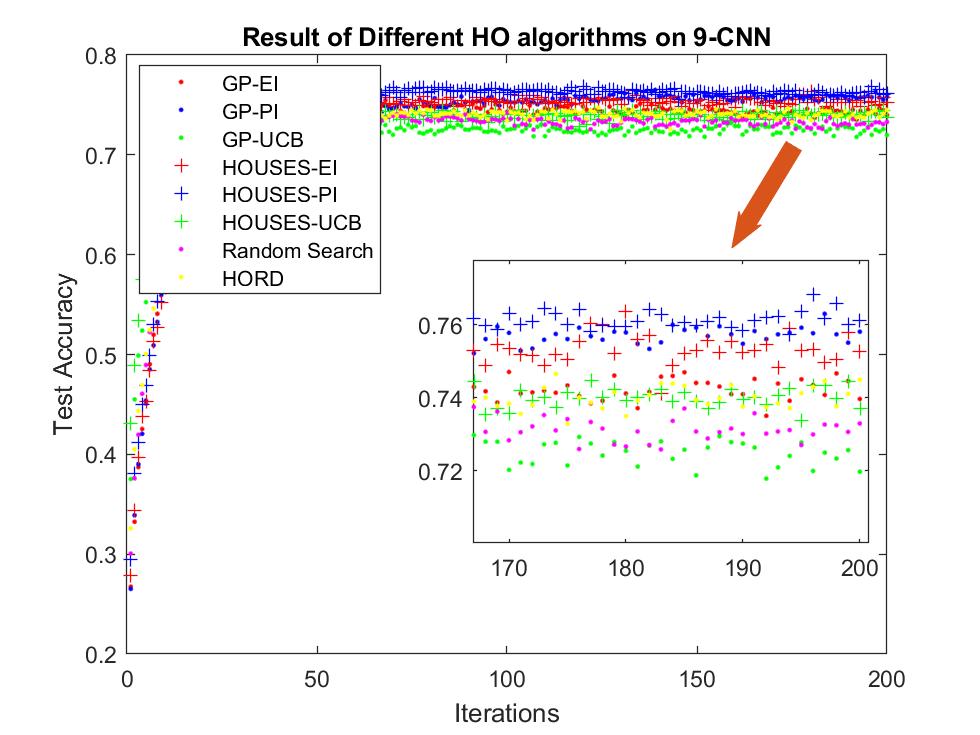}
      \caption*{(c)Testing accuracy of all hyperparameter optimization algorithms on 9-CNN problem.}
  \end{minipage}
  \begin{minipage}{6cm}
      \includegraphics[width=6cm]{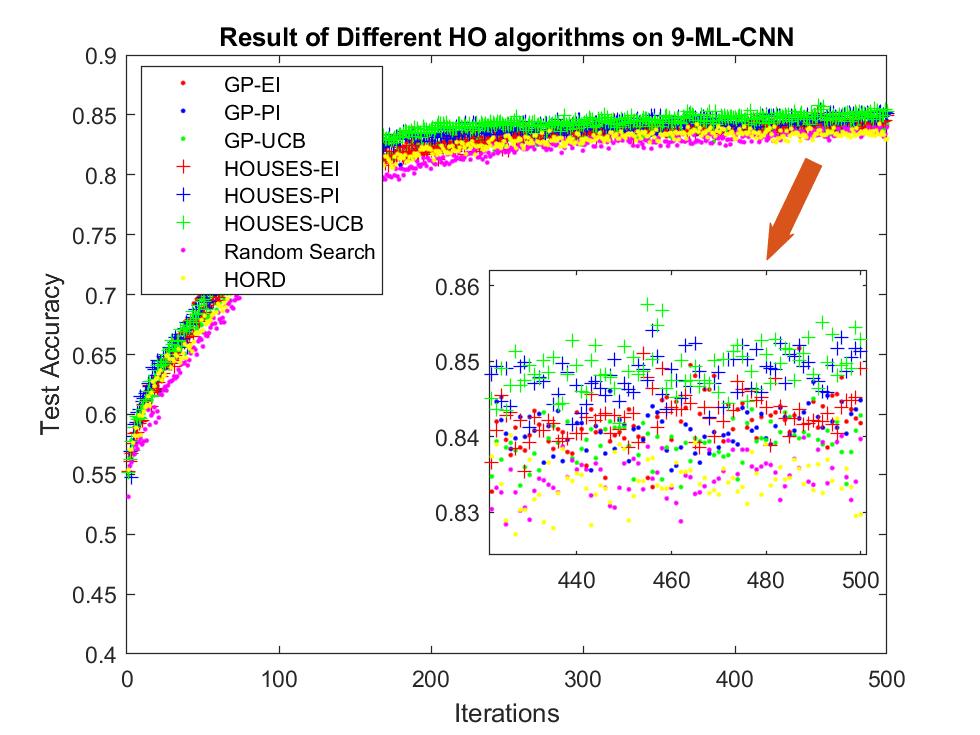}
      \caption*{\quad (d) Testing accuracy of all hyperparameter optimization algorithms on 9-ML-CNN problem.}
  \end{minipage}
  
  \caption{Testing accuracy of all hyperparameter optimization algorithms on four DNN problems.}
 \label{figure4}
\end{figure*}

\subsection{Experiments on LIDC-IDRI (Lung noddule classification problem)}

In this section, we evaluate HOUSES and all comparing algorithms applied to 9-ML-CNN (Multi-Level Convolutional Neural Network applied to lung nodule classification with 9 hyperparameters to be optimized), and the results are demomstrated in Table 1 Column 4, and Fig.\ref{figure4}. As expected, the performance of conventional hyperparameter optimization methods degrades significantly in complicated and high dimensional search space, while HOUSES continues to get satisfied results and outperforms the Gaussian process with stationary kernels. Similar to those result in previous subsection, we found that UCB gets the best result among three acquisition functions, which also suggests that UCB may be the most appropriate acquisition function in 9-ML-CNN hyperparameter optimization.

We presents the test accuracy over iterations of the obtained hyperparameter configurations for 9-ML-CNN problem from different hyperparameter optimization methods in Figure 2 (d). It is obvious that our spatial location based non-stationary kernel outperform stationary Gaussian kernel with three different acquisition functions, which also indicates a non-stationary kernel is especially necessary for complicated CNN hyperparameter optimization. We observe that HOUSES-UCB reaches better validation accuracy in only 250 epochs than manual tuning method \cite{lyvjuan}. Moreover, Table 3 present the ability of ML-CNN with hyperparameter configurations obtained by different hyperparameter optimization methods to classify different type of malignant nodules, and shows the sensitivity, specificity and AUC on three types of malignant nodules.

Results in 9-ML-CNN problem again shows that using a non-stationary kernel that takes the spatial location in consideration significantly improves the convergence of the hyperparameter optimization, especially for high-dimensional and complicated deep neural networks, and our hyperparameter optimization method HOUSES could not only relieve the trivial work to tune hyperparameters, but also get better results in terms of accuracy compared with manual tuning \cite{lyvjuan}.  Experimental results from above show that the non-stationary assumption is non-trivial for hyperparameter optimization in DNN with Bayesian methods, and incorporating priors based on expert intuition into Bayesian optimization framework is supposed to improve the optimization effectiveness.

\begin{sidewaystable}[!htbp]
\centering 
    \caption{Comparison results of 9-ML-CNN problem for each class}
    \label{t3}
    \begin{center}
        \begin{tabular}{c|ccc|ccc|ccc}
            \cline{1-10}
            \multirow{2}{*}{Algorithm} &\multicolumn{3}{c|}{Benign}&\multicolumn{3}{c|}{Indeterminate}&\multicolumn{3}{c}{Malignant}\\
            \cline{2-10}            
            &Sensitivity&Specificity&AUC&Sensitivity&Specificity&AUC&Sensitivity&Specificity&AUC\\
            \hline
            \hline
            Random Search&0.79260&0.92026&0.86643&0.83593&0.85256&0.85925&0.81320&0.92327&0.87814\\
            \hline
            HORD&0.83457&0.92735&0.88096&0.83447&0.90906&0.87226&0.86095&0.92947&0.89521\\
            \hline
            GP-EI&0.83395&0.92615&0.88005&0.83790&0.91208&0.87499&0.85685&0.92640&0.89160\\
            \hline
            GP-PI&0.81913&0.93726&0.89239&0.85314&0.91601&0.87604&0.85385&0.91874&0.88955\\
            \hline
            GP-UCB&0.82407&0.93155&0.88707&0.85314&\textbf{0.91994}&\textbf{0.87740}&0.85385&0.92364&0.89792\\
            \hline
            HOUSES-EI&0.84259&\textbf{0.94116}&0.88262&0.83406&0.89758&0.87536&0.87219&0.92702&0.89043\\
            \hline
            HOUSES-PI&0.84753&0.93966&0.87940&0.83608&0.89819&0.87109&0.86035&0.92180&0.88871\\
            \hline
            HOUSES-UCB&\textbf{0.85000}&0.93546&\textbf{0.89273}&0.82328&0.91516&0.86919&\textbf{0.87745}&0.92000&\textbf{0.89371}\\
            \hline   
            Manual Tuning&0.80617&0.93455&0.87036&\textbf{0.87751}&0.85468&0.86610&0.79882&\textbf{0.95216}&0.87549\\
            \hline   
        \end{tabular}
    \end{center}
\end{sidewaystable}

\section{Conclusion}

In this paper, a \textbf{H}yperparameter \textbf{O}ptimization with s\textbf{U}rrogate-a\textbf{S}sisted \textbf{E}volutionary \textbf{S}trategy, named HOUSES is proposed for CNN hyperparameter optimization. A non-stationary kernel is devised and adopted as covariance function to define the relationship between different hyperparameter configurations to build Gaussian processes model, which allows the model adapts spatial dependence structure to vary with a function of location. Our previous proposed multi-level convolutional neural network (ML-CNN) is developed for lung nodule malignancy classification, whose hyperparameter configuration is optimized by our HOUSES. Experimental results on several deep neural networks and datasers validated that our non-stationary kernel based approach could find better hyperparameter configuration than other approaches, such as Random search, Tree-structured Parzen Estimator (TPE), Hyperparameter Optimization via RBF and Dynamic coordinate search (HORD), and stationary kernel based Gaussian kernel Bayesian optimization.  Experimental results suggest that, even though Random Search is a simple and effective way for CNN hyperparameter optimization, it is hard to find satisfactory configuration for high-dimensional and complex deep neural networks, and incorporating priors based on expert intuition into conventional Bayesian optimization framework is supposed to improve the optimization effectiveness. Furthermore, the results also demonstrate devising a suitable network structure is a more robust way to improve performance, while hyperparameter optimization could help achieve the potential of the network. 

In light of the promising initial research results, our future research will focus on extending HOUSES to deep neural networks architecture search. Several works have been proposed to automatically search for well-performing CNN architectures via hill climbing procedure \cite{Elsken2017Simple}, Q-Learning \cite{Zhong2018Practical}, sequential model-based optimization (SMBO), and so on \cite{Negrinho2018DeepArchitect}, and genetic programming approach \cite{Suganuma2017A}. However, there are few works that utilize surrogate model to reduce the expensive complexity required by CNN searching. Moreover, a simple evolutionary strategy is not a appropriate method to search the surrogate for optimal architecture design, which is a variable-length optimization problem \cite{Lehman2011Evolving}, and quality-diversity based evolutionary algorithm may provide a solution to it.


\begin{thebibliography}{}
\expandafter\ifx\csname url\endcsname\relax
  \def\url#1{\texttt{#1}}\fi
\expandafter\ifx\csname urlprefix\endcsname\relax\def\urlprefix{URL }\fi
\expandafter\ifx\csname href\endcsname\relax
  \def\href#1#2{#2} \def\path#1{#1}\fi

\end{thebibliography}


\begin{thebibliography}{10}

\bibitem{3Rd2012The}
Armato~Sg 3Rd, G~Mclennan, L~Bidaut, M.~F. Mcnitt-Gray, C.~R. Meyer, A.~P.
  Reeves, B.~Zhao, D.~R. Aberle, C.~I. Henschke, and E.~A. Hoffman.
\newblock The lung image database consortium (lidc) and image database resource
  initiative (idri): a completed reference database of lung nodules on ct
  scans.
\newblock {\em Medical Physics}, 38(2):915, 2012.

\bibitem{Anthimopoulos2016Lung}
M~Anthimopoulos, S~Christodoulidis, L~Ebner, A~Christe, and S~Mougiakakou.
\newblock Lung pattern classification for interstitial lung diseases using a
  deep convolutional neural network.
\newblock {\em IEEE Transactions on Medical Imaging}, 35(5):1207--1216, 2016.

\bibitem{Bergstra2012Making}
J~Bergstra, D~Yamins, and D.~D Cox.
\newblock Making a science of model search.
\newblock 2012.

\bibitem{Bergstra2011Algorithms}
James Bergstra and Yoshua Bengio.
\newblock Algorithms for hyper-parameter optimization.
\newblock In {\em International Conference on Neural Information Processing
  Systems}, pages 2546--2554, 2011.

\bibitem{Bergstra2012Random}
James Bergstra and Yoshua Bengio.
\newblock Random search for hyper-parameter optimization.
\newblock {\em Journal of Machine Learning Research}, 13(1):281--305, 2012.

\bibitem{Chatzilygeroudis2017Black}
Konstantinos Chatzilygeroudis, Roberto Rama, Rituraj Kaushik, Dorian Goepp,
  Vassilis Vassiliades, and Jean~Baptiste Mouret.
\newblock Black-box data-efficient policy search for robotics.
\newblock In {\em Ieee/rsj International Conference on Intelligent Robots and
  Systems}, 2017.

\bibitem{Chien2003Pattern}
Y.~Chien.
\newblock Pattern classification and scene analysis.
\newblock {\em IEEE Transactions on Automatic Control}, 19(4):462--463, 2003.

\bibitem{Clark2013The}
Kenneth Clark, Bruce Vendt, Kirk Smith, John Freymann, Justin Kirby, Paul
  Koppel, Stephen Moore, Stanley Phillips, David Maffitt, and Michael Pringle.
\newblock The cancer imaging archive (tcia): Maintaining and operating a public
  information repository.
\newblock {\em Journal of Digital Imaging}, 26(6):1045--1057, 2013.

\bibitem{Dong2018A}
Hongbin Dong, Tao Li, Rui Ding, and Jing Sun.
\newblock A novel hybrid genetic algorithm with granular information for
  feature selection and optimization.
\newblock {\em Applied Soft Computing}, 65, 2018.

\bibitem{Douguet2010e}
Dominique Douguet.
\newblock e-lea3d: a computational-aided drug design web server.
\newblock {\em Nucleic Acids Research}, 38(Web Server issue):615--21, 2010.

\bibitem{Elsken2017Simple}
Thomas Elsken, Jan-Hendrik Metzen, and Frank Hutter.
\newblock Simple and efficient architecture search for convolutional neural
  networks.
\newblock {\em arXiv preprint arXiv:1711.04528}, 2017.

\bibitem{Genton2002Classes}
Marc~G. Genton.
\newblock Classes of kernels for machine learning: A statistics perspective.
\newblock {\em Journal of Machine Learning Research}, 2(2):299--312, 2002.

\bibitem{6909475}
R.~Girshick, J.~Donahue, T.~Darrell, and J.~Malik.
\newblock Rich feature hierarchies for accurate object detection and semantic
  segmentation.
\newblock In {\em 2014 IEEE Conference on Computer Vision and Pattern
  Recognition (CVPR)}, volume~00, pages 580--587, June 2014.

\bibitem{Golan2016Lung}
Rotem Golan, Christian Jacob, and Jörg Denzinger.
\newblock Lung nodule detection in ct images using deep convolutional neural
  networks.
\newblock In {\em International Joint Conference on Neural Networks}, pages
  243--250, 2016.

\bibitem{Hoffman2014Predictive}
Matthew~W. Hoffman and Zoubin Ghahramani.
\newblock Predictive entropy search for efficient global optimization of
  black-box functions.
\newblock In {\em International Conference on Neural Information Processing
  Systems}, pages 918--926, 2014.

\bibitem{hoos2014efficient}
Holger Hoos and Kevin Leyton-Brown.
\newblock An efficient approach for assessing hyperparameter importance.
\newblock In {\em International Conference on Machine Learning}, pages
  754--762, 2014.

\bibitem{Ilija}
Ilija Ilievski, Taimoor Akhtar, Jiashi Feng, and Christine Shoemaker.
\newblock Efficient hyperparameter optimization for deep learning algorithms
  using deterministic rbf surrogates, 2017.

\bibitem{Iman2008Latin}
Ronald~L. Iman.
\newblock {\em Latin Hypercube Sampling}.
\newblock John Wiley \& Sons, Ltd, 2008.

\bibitem{Jin2011Surrogate}
Yaochu Jin.
\newblock Surrogate-assisted evolutionary computation: Recent advances and
  future challenges.
\newblock {\em Swarm and Evolutionary Computation}, 1(2):61--70, 2011.

\bibitem{Jin2009A}
Yaochu Jin and B~Sendhoff.
\newblock A systems approach to evolutionary multiobjective structural
  optimization and beyond.
\newblock {\em Computational Intelligence Magazine IEEE}, 4(3):62--76, 2009.

\bibitem{Krizhevsky2012ImageNet}
Alex Krizhevsky, Ilya Sutskever, and Geoffrey~E. Hinton.
\newblock Imagenet classification with deep convolutional neural networks.
\newblock In {\em International Conference on Neural Information Processing
  Systems}, pages 1097--1105, 2012.

\bibitem{Lecun1998Efficient}
Yann Lecun, Leon Bottou, Genevieve~B. Orr, and Klaus~Robert Müller.
\newblock Efficient backprop.
\newblock {\em Neural Networks Tricks of the Trade}, 1524(1):9--50, 1998.

\bibitem{Lehman2011Evolving}
Joel Lehman and Kenneth~O Stanley.
\newblock Evolving a diversity of virtual creatures through novelty search and
  local competition.
\newblock In {\em Proceedings of the 13th annual conference on Genetic and
  evolutionary computation}, pages 211--218. ACM, 2011.

\bibitem{Liu2017Multiview}
Kui Liu and Guixia Kang.
\newblock Multiview convolutional neural networks for lung nodule
  classification.
\newblock {\em Plos One}, 12(11):12--22, 2017.

\bibitem{DBLP:journals/corr/LoshchilovH16}
Ilya Loshchilov and Frank Hutter.
\newblock {CMA-ES} for hyperparameter optimization of deep neural networks.
\newblock {\em CoRR}, abs/1604.07269, 2016.

\bibitem{lyvjuan}
Juan Lyu and Sai~Ho Ling.
\newblock Using multi-level convolutional neural network for classification of
  lung nodules on ct images.
\newblock In {\em 2018 40th Annual International Conference of the IEEE
  Engineering in Medicine and Biology Society (EMBC)}, pages 686--689. IEEE,
  2018.

\bibitem{Min2017Multi}
Alan Tan~Wei Min, Yew~Soon Ong, Abhishek Gupta, and Chi~Keong Goh.
\newblock Multi-problem surrogates: Transfer evolutionary multiobjective
  optimization of computationally expensive problems.
\newblock {\em IEEE Transactions on Evolutionary Computation}, PP(99):1--1,
  2017.

\bibitem{Negrinho2017DeepArchitect}
Renato Negrinho and Geoff Gordon.
\newblock Deeparchitect: Automatically designing and training deep
  architectures.
\newblock 2017.

\bibitem{Negrinho2018DeepArchitect}
Renato Negrinho and Geoff Gordon.
\newblock Deeparchitect: Automatically designing and training deep
  architectures.
\newblock {\em arXiv preprint arXiv:1704.08792}, 2017.

\bibitem{LIDC}
Anthony~P. Reeves and Alberto~M. Biancardi.
\newblock The lung image database consortium (lidc) nodule size report.
\newblock \url{http://www.via.cornell.edu/lidc/}, 20011.

\bibitem{Reeves2007The}
Anthony~P. Reeves, Alberto~M. Biancardi, Tatiyana~V. Apanasovich, Charles~R.
  Meyer, Heber Macmahon, Edwin J. R.~Van Beek, Ella~A. Kazerooni, David
  Yankelevitz, Michael~F. Mcnittgray, and Geoffrey Mclennan.
\newblock The lung image database consortium (lidc): pulmonary nodule
  measurements, the variation, and the difference between different size
  metrics.
\newblock In {\em Medical Imaging 2007: Computer-Aided Diagnosis}, pages
  1475--1485, 2007.

\bibitem{Regis2013Combining}
Rommel~G. Regis and Christine~A. Shoemaker.
\newblock Combining radial basis function surrogates and dynamic coordinate
  search in high-dimensional expensive black-box optimization.
\newblock {\em Engineering Optimization}, 45(5):529--555, 2013.

\bibitem{Shahriari2015Taking}
Bobak Shahriari, Kevin Swersky, Ziyu Wang, Ryan~P. Adams, and Nando~De Freitas.
\newblock Taking the human out of the loop: A review of bayesian optimization.
\newblock {\em Proceedings of the IEEE}, 104(1):148--175, 2015.

\bibitem{Shen2015Multi}
W.~Shen, M.~Zhou, F.~Yang, C.~Yang, and J.~Tian.
\newblock Multi-scale convolutional neural networks for lung nodule
  classification.
\newblock {\em Inf Process Med Imaging}, 24:588--599, 2015.

\bibitem{Shen2017Multi}
Wei Shen, Mu~Zhou, Feng Yang, Dongdong Yu, Di~Dong, Caiyun Yang, Yali Zang, and
  Jie Tian.
\newblock Multi-crop convolutional neural networks for lung nodule malignancy
  suspiciousness classification.
\newblock {\em Pattern Recognition}, 61(61):663--673, 2017.

\bibitem{Siegel2017Colorectal}
R.~L. Siegel, K.~D. Miller, S.~A. Fedewa, D.~J. Ahnen, R.~G. Meester, A~Barzi,
  and A~Jemal.
\newblock Colorectal cancer statistics, 2017.
\newblock {\em Ca Cancer J Clin}, 67(3):104--17, 2017.

\bibitem{Snoek2012Practical}
Jasper Snoek, Hugo Larochelle, and Ryan~P. Adams.
\newblock Practical bayesian optimization of machine learning algorithms.
\newblock In {\em International Conference on Neural Information Processing
  Systems}, pages 2951--2959, 2012.

\bibitem{snoek2014input}
Jasper Snoek, Kevin Swersky, Rich Zemel, and Ryan Adams.
\newblock Input warping for bayesian optimization of non-stationary functions.
\newblock In {\em International Conference on Machine Learning}, pages
  1674--1682, 2014.

\bibitem{Song2017Using}
Q.~Song, L.~Zhao, X.~Luo, and X.~Dou.
\newblock Using deep learning for classification of lung nodules on computed
  tomography images.
\newblock {\em J Healthc Eng.}, 2017(1):1--7, 2017.

\bibitem{Suganuma2017A}
Masanori Suganuma, Shinichi Shirakawa, and Tomoharu Nagao.
\newblock A genetic programming approach to designing convolutional neural
  network architectures.
\newblock pages 497--504, 2017.

\bibitem{Sun2016Computer}
Wenqing Sun, Bin Zheng, and Wei Qian.
\newblock Computer aided lung cancer diagnosis with deep learning algorithms.
\newblock In {\em Medical Imaging 2016: Computer-Aided Diagnosis}, 2016.

\bibitem{swersky2017improving}
Kevin~Jordan Swersky.
\newblock {\em Improving Bayesian Optimization for Machine Learning using
  Expert Priors}.
\newblock PhD thesis, 2017.

\bibitem{Naiyan2015Transferring}
Naiyan Wang, Siyi Li, Abhinav Gupta, and DitYan Yeung.
\newblock Transferring rich feature hierarchies for robust visual tracking.
\newblock {\em Computer Science}, 2015.

\bibitem{Zhang2018A}
Miao Zhang and Huiqi Li.
\newblock A reference direction and entropy based evolutionary algorithm for
  many-objective optimization.
\newblock {\em Applied Soft Computing}, 70:108--130, 2018.

\bibitem{Zhang2010Expensive}
Qingfu Zhang, Wudong Liu, Edward Tsang, and Botond Virginas.
\newblock Expensive multiobjective optimization by moea/d with gaussian process
  model.
\newblock {\em IEEE Transactions on Evolutionary Computation}, 14(3):456--474,
  2010.

\bibitem{Zhong2018Practical}
Zhao Zhong, Junjie Yan, Wei Wu, Jing Shao, and Cheng-Lin Liu.
\newblock Practical block-wise neural network architecture generation.
\newblock pages 2423--2432, 2018.

\end{thebibliography}
\end{document}